\def\year{2021}\relax
\documentclass[letterpaper]{article} \usepackage{aaai21}  \usepackage{times}  \usepackage{helvet} \usepackage{courier}  \usepackage[hyphens]{url}  \usepackage{graphicx}    \usepackage{natbib}  \usepackage{caption} \usepackage{amsmath}
\usepackage{amssymb}
\usepackage{multirow}
\usepackage{amsthm}
\usepackage{soul, color}
\usepackage{xcolor}
\usepackage{multirow}
\usepackage{tabularx}
\usepackage{framed}
\usepackage{booktabs} 
\title{Neural Sequence-to-grid Module for Learning Symbolic Rules}
\author{
Segwang Kim, \textsuperscript{\rm 1} Hyoungwook Nam, \textsuperscript{\rm 2} Joonyoung Kim, \textsuperscript{\rm 1} Kyomin Jung\textsuperscript{\rm 1}
}

\affiliations{

    \textsuperscript{\rm 1} Department of Electrical and Computer Engineering, Seoul National University, Seoul, Republic of Korea\\
    \textsuperscript{\rm 2} Department of Computer Science, University of Illinois at Urbana-Champaign, Urbana, Illinois, USA\\

$\{$ksk5693, kimjymcl, kjung$\}$@snu.ac.kr, hn5@illinois.edu 

}

\begin{document}

\maketitle

\begin{abstract}
Logical reasoning tasks over symbols, such as learning arithmetic operations and computer program evaluations, have become challenges to deep learning.
In particular, even state-of-the-art neural networks fail to achieve \textit{out-of-distribution} (OOD) generalization of symbolic reasoning tasks, whereas humans can easily extend learned symbolic rules.
To resolve this difficulty, we propose a neural sequence-to-grid (seq2grid) module, an input preprocessor that automatically segments and aligns an input sequence into a grid.
As our module outputs a grid via a novel differentiable mapping, any neural network structure taking a grid input, such as ResNet or TextCNN, can be jointly trained with our module in an end-to-end fashion.
Extensive experiments show that neural networks having our module as an input preprocessor achieve OOD generalization on various arithmetic and algorithmic problems including number sequence prediction problems, algebraic word problems, and computer program evaluation problems while other state-of-the-art sequence transduction models cannot.
Moreover, we verify that our module enhances TextCNN to solve the bAbI QA tasks without external memory.
 \end{abstract}
 
\section{Introduction}
\noindent Symbolic reasoning tasks such as learning arithmetic operations or evaluating computer programs offer solid standards for validating the logical inference abilities of deep learning models.
Among machine learning tasks, symbolic reasoning problems are apt for testing mathematical, algorithmic, and systematic reasoning as they have strict rules mapping a given input to a well-defined unique target.
In particular, a large body of works on deep learning has considered sequence transduction problems for symbolic reasoning.
Some symbolic problems such as copying sequences \cite{dehghani2018universal, graves2014neural, grefenstette2015learning, rae2016scaling, zaremba2014learning} and arithmetic addition \cite{graves2014neural, joulin2015inferring, kaiser2015neural, kalchbrenner2015grid, saxton2019analysing, wangperawong2018attending} can be solved after understanding simple rules regardless of the inputs.
Others demand a deep learning model to discover necessary rules and apply them depending on inputs given as natural language words \cite{li2019modeling, wang2017deep, weston2015towards}, complex mathematical equations \cite{lample2019deep}, or programming snippets \cite{zaremba2014learning}.

Among them, symbolic reasoning problems can test whether a trained deep learning model can systematically extend rules to \textit{out-of-distribution} (OOD) data that follow a distinct distribution from the training data \cite{keysers2019measuring, lake2017generalization, saxton2019analysing}.
For instance, a model for the addition problem whose training inputs are a pair of numbers up to five digits, say \texttt{5872+13}, can face an OOD input of a pair of two 6-digit numbers upon the testing phase, e.g., \texttt{641436+135321}.
Human intelligence with \textit{algebraic mind} can naturally extend learned rules \cite{marcus2001algebraic}, yet it is non-trivial to equip deep learning models for sequence transduction problems to handle OOD generalization.

However, it has been found that popular sequence transduction neural networks, such as LSTM seq2seq model \cite{sutskever2014sequence} and Transformer \cite{vaswani2017attention}, rarely extend learned rules in that they are inclined to mimic the training data distribution \cite{dehghani2018universal, lake2017generalization}.
There have been significant initial efforts to improve a model's abilities to extend learned rules.
However, their success has been dependent on the direct use of numerical values \cite{trask2018neural} or has been limited to rudimentary logic such as copying sequences \cite{dehghani2018universal, graves2014neural, grefenstette2015learning, rae2016scaling, zaremba2014learning} and binary arithmetic \cite{graves2014neural, joulin2015inferring, kaiser2015neural}.
Furthermore, OOD generalization on symbolic problems for complex or context-dependent logic forms such as decimal arithmetic, algebraic word problems, computer program evaluation problems has not been tackled.
Our objective is to fill this gap and design a module that helps neural networks to achieve OOD generalization in these problems.

One observation from a previous study \cite{nam2019number} is that typical sequence transduction neural networks cannot process OOD instances of number sequence prediction problems, such as predicting a Fibonacci sequence.
However, when an input sequence is manually segmented and aligned into a grid of digits, a CNN can easily process OOD instances.
This means providing the aligned grid input enables to exploit inductive bias by the convolution's local and parallel computation.
The grid, however, must be handcrafted in the study, which is inapplicable for general sequence transduction tasks.
Overcoming this limitation requires a new input preprocessing module that automatically aligns an input sequence into a grid without supervision for the alignment.

In this work, we propose a neural sequence-to-grid (seq2grid) module, an input preprocessor that learns how to segment and align an input sequence into a grid.
The grid syntactically aligned by our module is then semantically decoded by a neural network.
In particular, our module produces a grid by a novel differentiable mapping called nested list operation inspired by Stack RNN \cite{joulin2015inferring}.
This mapping enables a joint training of our module and the neural network in an end-to-end fashion via a backpropagation.

Experimental results show that ResNets with our seq2grid module achieve OOD generalization on various arithmetic and algorithmic reasoning problems, such as number sequence prediction problems, algebraic word problems, and computer program evaluation problems.
These are nearly impossible for other contemporary sequence-to-sequence models including LSTM seq2seq models and Transformer-based models.
Specifically, we find that the seq2grid can infuse an input context into a grid so that doing arithmetic under linguistic instructions or selecting the true branch of if/else statements in code snippets become possible.
Further, we demonstrate that the seq2grid module can enhance TextCNN to solve the bAbI QA tasks without the help of external memory.
From all the aforementioned problems, we verify the generality of the seq2grid module in that it automatically preprocesses the sequential input into the grid input in a data-driven way.

\section{Motivation for Sequence-to-grid Method}
\label{section:sequence_alignment_over_grid}
\begin{figure}[ht]
\centerline{\includegraphics[width=0.82\columnwidth]{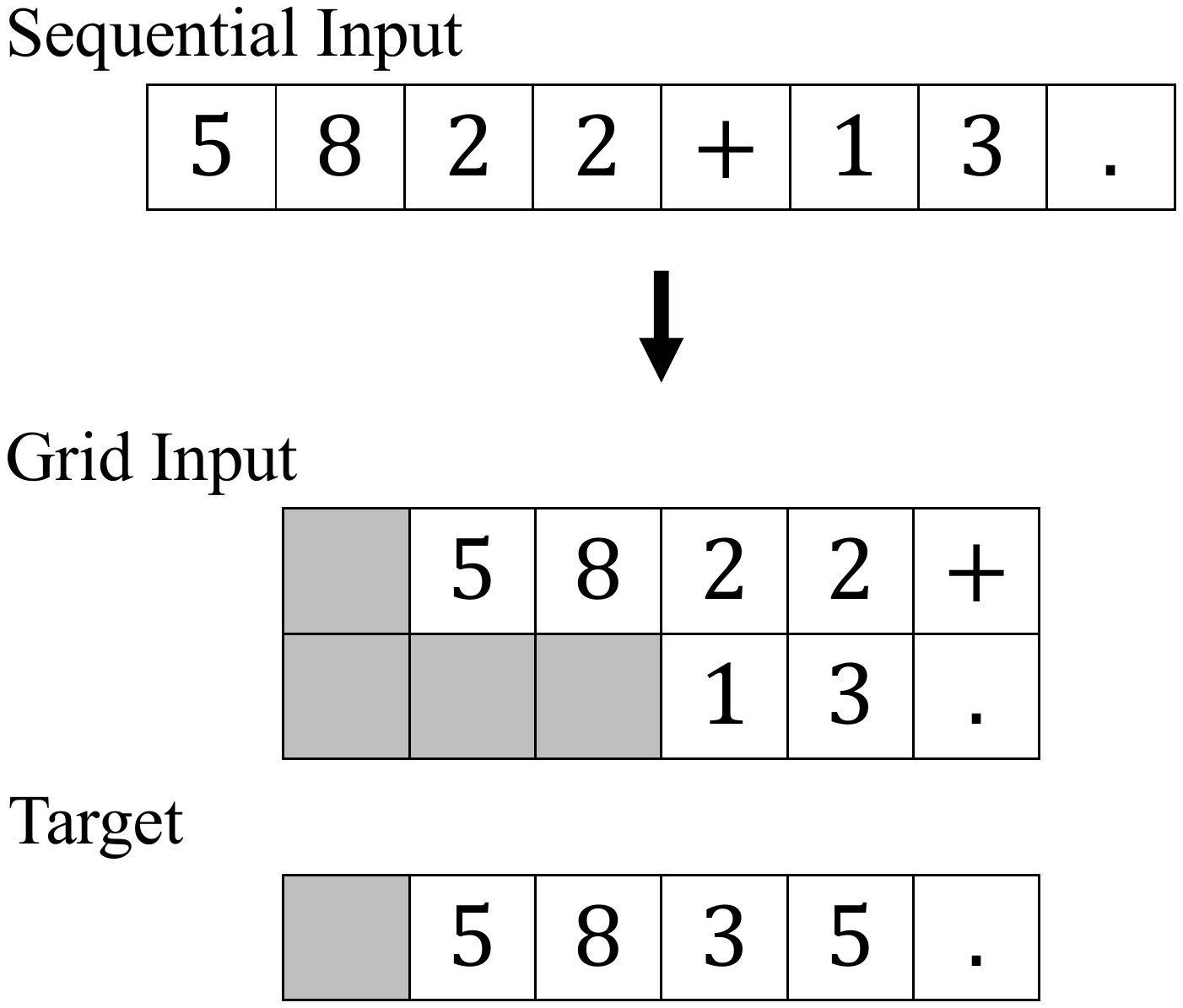}}
\caption{The illustration of the toy decimal addition problem. Each symbol is stored with its representation vector.} \label{fig:toy_a}
\vskip 0.3in 
\centerline{\includegraphics[width=0.9\columnwidth]{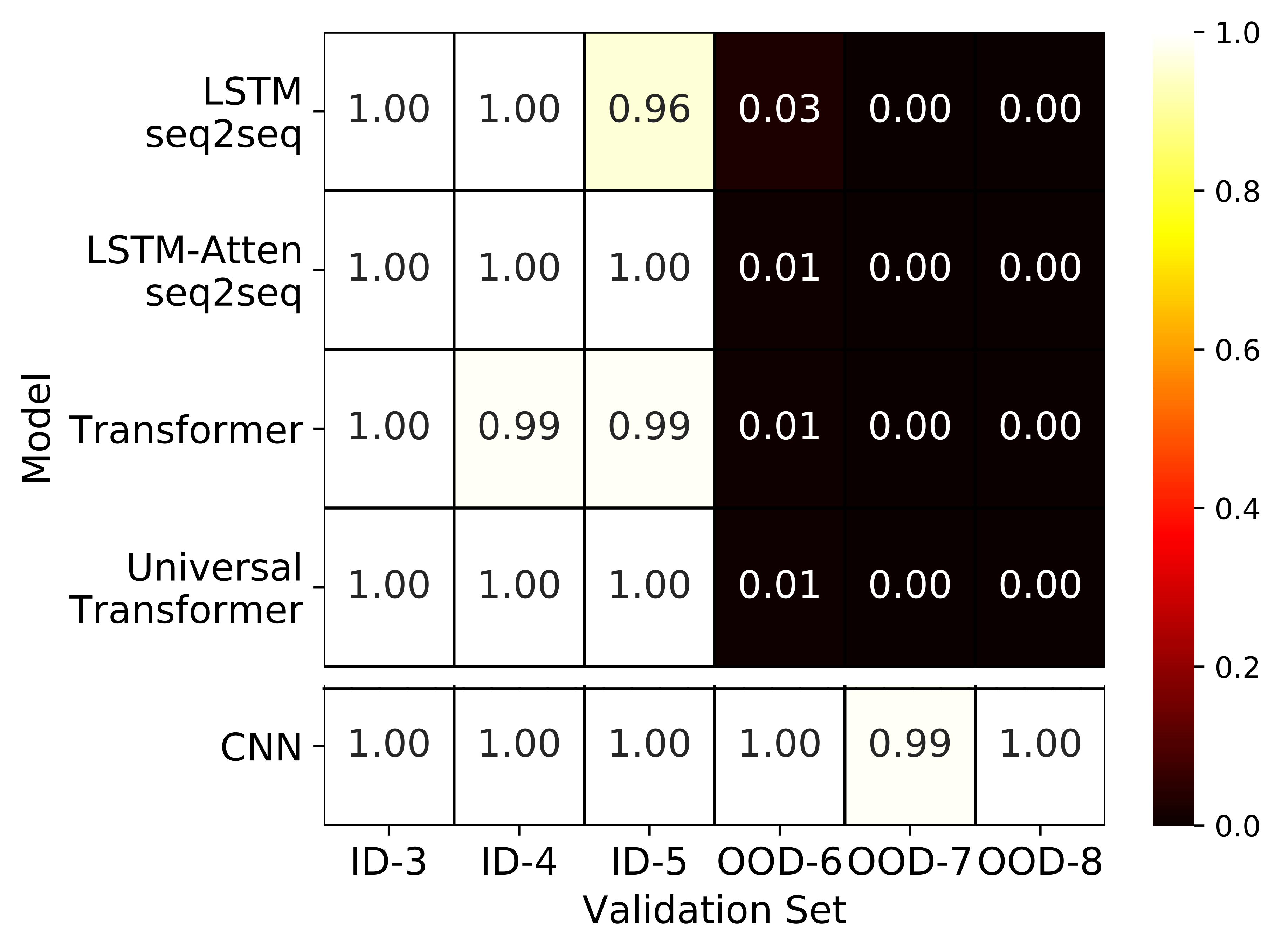}}
\caption{The validation accuracy results of the toy problem. 
Each column shows results from the $k$-digit set, where the three rightmost sets are OOD.} \label{fig:toy_b}
\end{figure}

To demonstrate the benefits of the sequence-to-grid preprocessing method for symbolic reasoning tasks, we devise a toy decimal addition problem in two different setups: sequential and grid-structured.
Figure~\ref{fig:toy_a} illustrates how the problem is defined in both setups and shows why alignment on a grid makes it easier.
If the lengths of the numbers increase, the temporal distances between corresponding digits, e.g., \texttt{2} and \texttt{3}, also increase in the sequential setup. 
However, the spatial distances between them remain constant in the grid-structured setup since they are \textit{manually} aligned according to their digits.
Therefore, we can expect that the local and parallel nature of convolution will extend the rule to longer inputs, while sequence transduction models will struggle to handle the increased distances.

To see this, we trained deep learning models\footnote{The models had the same configurations used in arithmetic and algorithmic problems (refer to experiments) except for the CNN that was the grid decoder of the S2G-CNN.} using numbers up to five digits and validate on six separate validation sets, each of which contains only $k$-digit ($k\!=\!3,\dots, 8$) numbers.
Hence, the validation results from the former three sets tested \textit{in-distribution} (ID) generalization, whereas the latter three tested OOD generalization.
While the input and the target in the sequential setup were sequentially fed to sequence transduction models such as LSTM seq2seq model \cite{sutskever2014sequence} and Transformer
\cite{vaswani2017attention}, those in the grid-structured setup were fed to a ResNet-based CNN model \cite{he2016deep}.
As expected, Figure~\ref{fig:toy_b} shows that extending the addition rule to OOD validation sets is easy in the grid-structured setup, whereas it is extremely difficult in the sequential setup.

Therefore, providing aligned grid input for local and parallel computation can be key to achieving OOD generalization.
However, manual preprocessing that aligns an input sequence into a grid is impossible for most symbolic problems.
For instance, in computer program evaluation problems, symbols within the code snippet can represent not only integers but also programming instructions so that it is non-trivial to manually align those symbols on the grid.
Likewise, in the bAbI QA tasks, questions and stories given as natural language have no ground-truth alignment which we can exploit for preprocessing in advance.
Accordingly, we need a data-driven preprocessing method that automatically aligns an input symbol sequence into a grid for general symbolic tasks.
We implement it by designing a sequence-to-grid module executing novel \textit{nested list operations}.

\section{Related Work}
\label{section:related_work}

\paragraph{Symbolic Reasoning Tasks}
Symbolic reasoning requires discovering the underlying rules of a data distribution rather than mimicking data patterns.
Hence, there have been studies to formulate symbolic reasoning tasks in machine learning problems to examine the mathematical and systematic (rule-based) reasoning abilities of deep learning models.
Additions of unprecedented long binary numbers \cite{graves2014neural, joulin2015inferring, kaiser2015neural} or Number sequence prediction problems \cite{nam2019number} are studied.
Also, school-level math problems including algebraic word problems are unified \cite{saxton2019analysing}.
Evaluating program snippets \cite{zaremba2014learning} further requires algorithmic abilities.
Besides extending mathematical rules, systematic generalization abilities in synthetic natural language tasks are tested by the bAbI QA tasks \cite{weston2015towards} or the SCAN problems \cite{lake2017generalization}.

\paragraph{Memory Augmented Neural Network}
Storing all input information into external memory and querying over it is one way to tackle symbolic reasoning tasks.
Such neural networks, also known as memory augmented neural networks (MANN), vary according to their memory structures and controllers.
Here, the memory controller is a neural network that reads an input symbol and its external memory, encodes the symbol, and write it on the memory.
After the incipient MANNs like Memory network \cite{sukhbaatar2015end} were introduced, studies about implementing Automata with differentiable tape as neural networks \cite{graves2016hybrid, rae2016scaling, joulin2015inferring, grefenstette2015learning} has been carried out.
We emphasize that our preprocessed grid input can be seen as another representation of a sequential input rather than a memory used in MANNs; 
the RNN encoder of our module does not read the grid and the symbol embedding is directly written to the grid rather than passed through neural network layers.

\paragraph{Neural-Symbolic Learning}
Another approach for OOD generalization in symbolic problems is Neural-Symbolic-approach that integrates the connectionist and the symbolist paradigms. 
Neural Programmer Interpreter (NPI) \cite{reed2015neural} and its recursion variant \cite{cai2017making} have been proposed for solving compositional programs through sequential subprograms. 
Also, \cite{chen2017towards} proposes a reinforcement learning-based approach with structured parse-trees. 
Recently, Neural Symbolic Reader \cite{chen2019neural} trains models with weak supervision for generalization.  
However, our approach via automatic alignment without domain-specific knowledge is distinct from neural-symbolic approaches which require all of the supervision for sequential sub-operations.  
\section{Method}
In this section, we first describe a sequence-input grid-output architecture consisting of a neural sequence-to-grid (seq2grid) module and a grid decoder.
Then, we introduce how the seq2grid module preprocesses a sequence input as a grid input.
Finally, we explain nested list operations that are executed by the seq2grid module.

\subsection{Sequence-input Grid-output Architecture}
\begin{figure}[t]
\centerline{\includegraphics[width=\columnwidth]{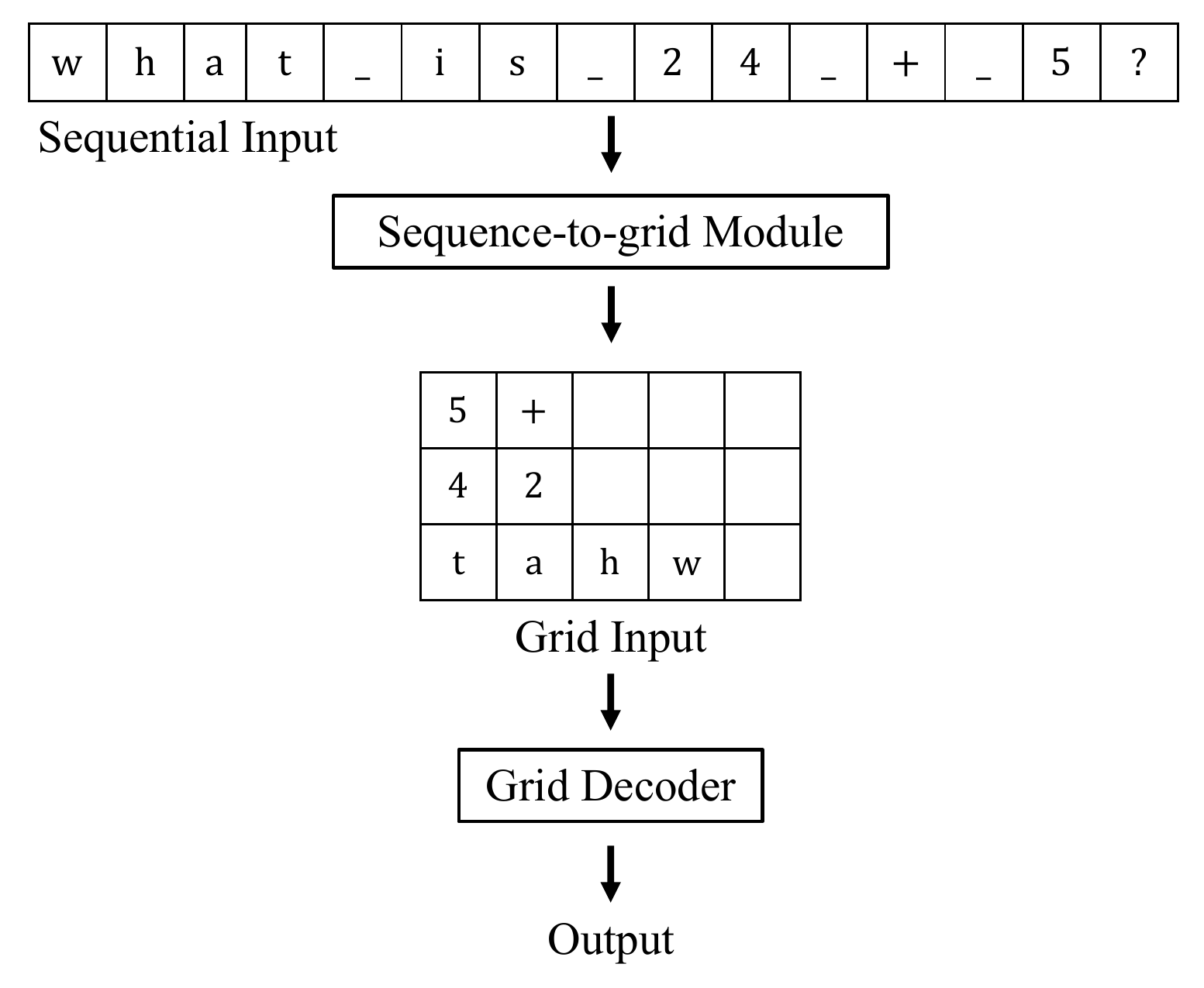}}
\caption{
The sequence-input grid-output architecture.}
\label{fig:SIGO_architecture}
\end{figure}

The key idea of the sequence-to-grid method is to decouple symbolic reasoning into two steps: automatically aligning an input sequence into a grid, and doing semantic computations over the grid.
Hence, we propose the sequence-input grid-output architecture consisting of a seq2grid module and a grid decoder as shown in Figure~\ref{fig:SIGO_architecture}.
The seq2grid module preprocesses a sequential input into a grid input.
The grid decoder, a neural network that can handle two-dimensional inputs, predicts the target from the grid input.
Practically, we choose the grid decoder like ResNet or TextCNN according to problems.
Note that our approach that separates the syntactic (=alignment) and semantic processing is similar to the syntactic attention \cite{russin2020compositional}.

\subsection{Neural Sequence-to-grid Module}
The main challenge for implementing the seq2grid module is that the grid must be formed via differentiable mappings to ensure an end-to-end training.
To do so, we design the seq2grid module with an RNN encoder that gives an action sequence for differentiable nested list operations.

Formally, the seq2grid module works as follows.
First, for an input sequence given as symbol embeddings $E^{(t)}\in \mathbb{R}^h$ where $t=1,\dots,T$, the RNN encoder maps $(E^{(t)}, r^{(t-1)})$ into $r^{(t)} \in \mathbb{R}^h$.
Then, a dense layer followed by a softmax layer computes an action: $r^{(t)} \mapsto a^{(t)} \in \mathbb{R}^3$.
Next, starting from the zero-initialized grid $G^{(0)} \in (\mathbb{R}^h)^{W {\times} H}$, a series of nested list operations sequentially push the input symbol $E^{(t)}$ into the previous grid $G^{(t-1)}$ in different extents under the action $a^{(t)}$.
As a result, we obtain the grid input $G^{(T)} \in (\mathbb{R}^h)^{W {\times} H}$ that will be fed through a grid decoder.
Note that all aforementioned mappings are differentiable including nested list operations which we will explain below.

\subsection{Nested List Operations}
To understand how the nested list operations work, we first regard the grid ${G} \in (\mathbb{R}^{h})^{H \times W}$ as a \textit{nested list} consisting of $H$ lists of $W$ slots, where each slot is a vector of dimension $h$.
We denote the $i$-th list as ${G}_i \in (\mathbb{R}^{h})^{W}$ where ${G}_1$ is the top list.
Likewise, the $j$-th slot vector in the $i$-th list is denoted as ${G}_{i,j} \in \mathbb{R}^h$ where ${G}_{i,1}$ is the leftmost slot of the $i$-th list.

\begin{figure}[t]
\centerline{\includegraphics[width=\columnwidth]{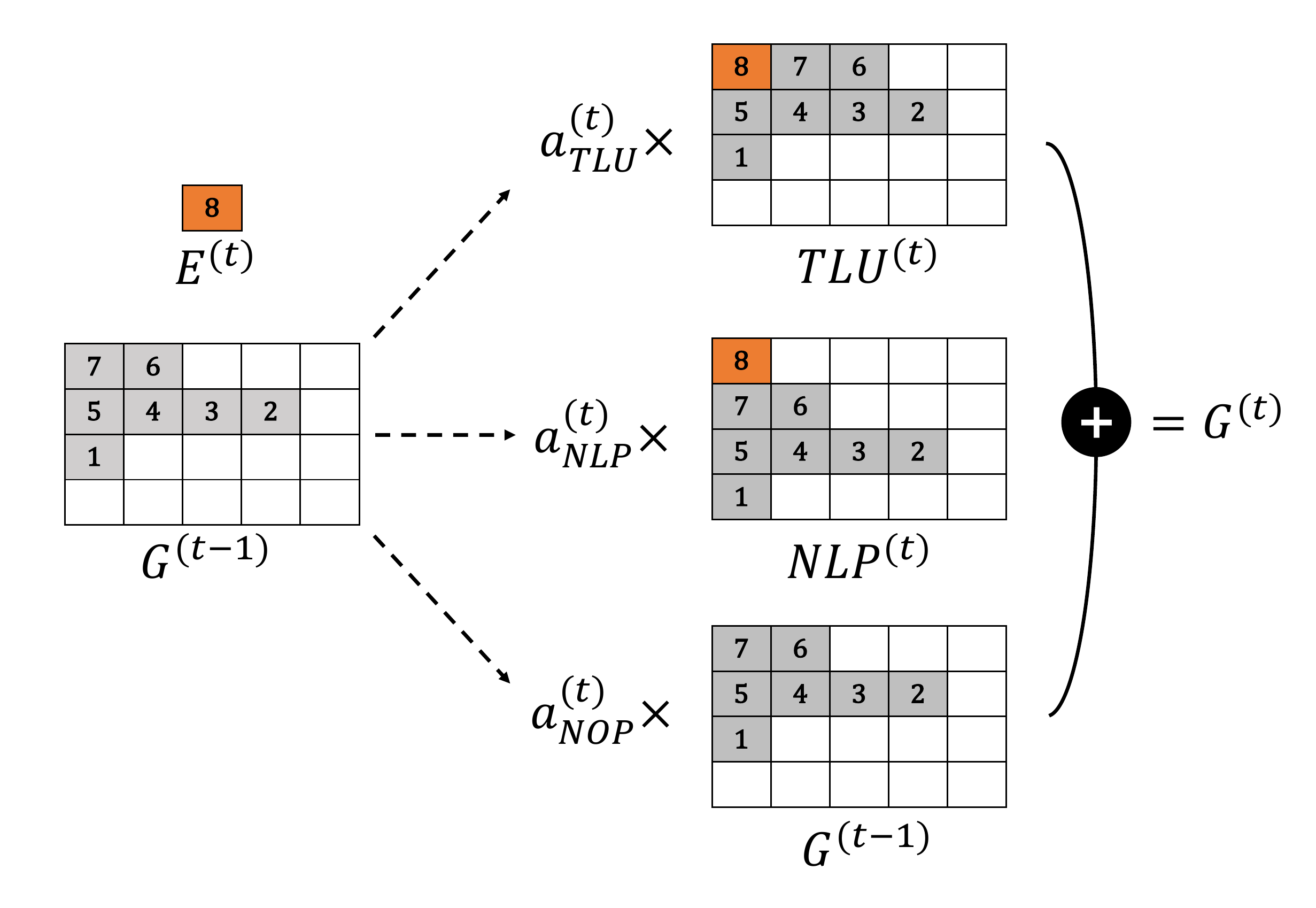}}
\caption{
The nested list $G^{(t-1)}$ grows to $G^{(t)}$ by the action $a^{(t)}=(a^{(t)}_{TLU}, a^{(t)}_{NLP}, a^{(t)}_{NOP})$.
$TLU^{(t)}$ and $NLP^{(t)}$ show outputs of \textit{top-list-update} and \textit{new-list-push} operations. 
}
\label{fig:nested_list_evolution}
\end{figure}

Now, we define a differetiable map that pushes the input symbol $E^{(t)} \in \mathbb{R}^h$ into the grid under the action $a^{(t)} \in \mathbb{R}^3$.
Here, each component of $a^{(t)} = (a^{(t)}_{TLU}, a^{(t)}_{NLP}, a^{(t)}_{NOP})$ is the probability of performing one of three nested list operations: \textit{top-list-update} ($a^{(t)}_{TLU}$), \textit{new-list-push} ($a^{(t)}_{NLP}$), and \textit{no-op} ($a^{(t)}_{NOP}$). 
As shown in Figure \ref{fig:nested_list_evolution}, ${G}^{(t-1)}$ with $(E^{(t)}, a^{(t)})$ grows to ${G}^{(t)}$:
\begin{align*}
{G}^{(t)} =  a^{(t)}_{TLU} {TLU}^{(t)} + a^{(t)}_{NLP} {NLP}^{(t)} + a^{(t)}_{NOP} {G}^{(t-1)},
\end{align*}
\begin{flalign*}
\mbox{where }{TLU}^{(t)} \in (\mathbb{R}^h&)^{H \times W}\mbox{ is defined as} &\\
    {TLU}^{(t)}_{1,1} &= E^{(t)},  \\
    {TLU}^{(t)}_{1,j} &= {G}^{(t-1)}_{1,j-1} \qquad \text{for } j > 1, \\
    {TLU}^{(t)}_i &= {G}^{(t-1)}_i \qquad \text{for } i > 1, \\
\mbox{and }{NLP}^{(t)} \in (\mathbb{R}^h&)^{H \times W}\mbox{ is defined }\mbox{as} &\\
{NLP}^{(t)}_1 &= (E^{(t)}, E_\emptyset, \dots , E_\emptyset),  \\
    {NLP}^{(t)}_i &= {G}^{(t-1)}_{i-1} \qquad \text{for } i > 1.
\end{flalign*}
Here, $E_\emptyset:=\boldsymbol{0}\in\mathbb{R}^h$ is the empty symbol $\emptyset$.
Accordingly, the zero-initialized grid $G^{(0)}=(E_\emptyset, \dots , E_\emptyset)$ grows to the final grid $G^{(T)}$ as time goes.
By doing so, we ``preprocess'' the input sequence into the grid input in that each slot of $G^{(T)}$ holds nothing but a weighted sum of input symbols.

\section{Experimental Setup}
\label{section:problems}
We evaluated the seq2grid module on symbolic problems whose targets are given as sequences or single labels.
To this end, we built neural network models, such as S2G-CNN and S2G-TextCNN, that followed the sequence-input grid-output architecture by varying the grid decoder according to target modalities of problems.
Refer to each problem section for our grid decoder choices and their training losses.

We compared our models with five baselines: Transformer \cite{vaswani2017attention}, Universal Transformer (UT) \cite{dehghani2018universal} with dynamic halting\footnote{The UT can take different ponder time for each position.}, a LSTM seq2seq model (LSTM) \cite{sutskever2014sequence}, a LSTM seq2seq attention model with a bidirectional encoder (LSTM-Atten) \cite{bahdanau2014neural} and a Relational Memory Core seq2seq model (RMC) \cite{santoro2018relational}. 
The Transformer and the UT consisted of two layers with the hidden size 128 and four attention heads.
The LSTM, the LSTM-Atten, and the RMC had three layers with the hidden size 1024, 512, and 512 each.

We determined configurations of our models by hyperparameter sweeping for each problem.
Our implementations\footnote{\label{our_code}https://github.com/SegwangKim/neural-seq2grid-module} based on the open source library \texttt{tensor2tensor}\footnote{https://github.com/tensorflow/tensor2tensor} contain detailed training schemes and hyperparameters of our models and the baselines.
All models could fit in a single NVIDIA GTX 1080ti GPU.

The next three sections follow the same organization to illustrate the experiments and their results.
First, we introduce a set of symbolic reasoning tasks by describing the layouts of the inputs and the targets.
Next, we describe our grid decoder architecture to solve such problems, which is combined with the seq2grid module to follow the sequence-input grid-output architecture.
Finally, we analyze the experimental results and discuss their implications. 
\section{Arithmetic and Algorithmic Problems}
\label{section:arithmetic_and_algorithmic}
\begin{figure}[t]
\centerline{\includegraphics[width=\columnwidth]{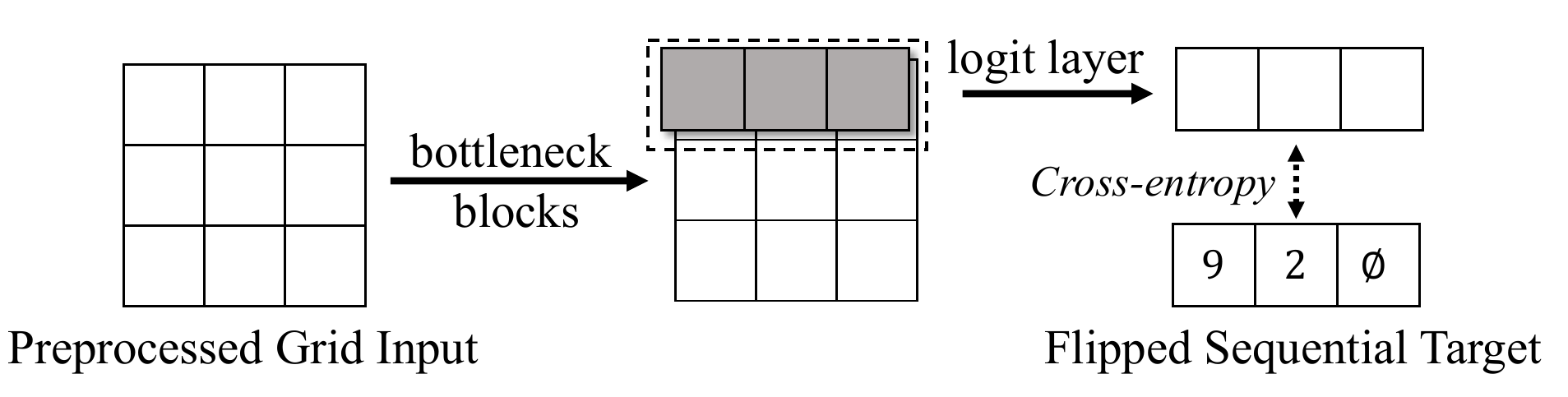}}
\caption{
The grid decoder of S2G-CNN.
Only the top list of the grid from bottleneck blocks is passed to the logit layer.
The raw target \texttt{29} is flipped and padded to $\texttt{92$\emptyset$}$.
}
\label{fig:S2G-CNN}
\end{figure}
Arithmetic and algorithmic problems are useful to test abilities to extend rules on longer inputs since the input contains digits.
We test our models on three different arithmetic and algorithmic inference problems.
Each problem consists of a training set and two test sets randomly sampled from distributions controlled by difficulty parameters.
Two test sets represent in-distribution data (ID) and OOD data (OOD).
Difficulty parameters of the training set can be overlapped with those of the ID test set, but instances of the two sets are strictly separated by their hash codes.
The training set of all problems contains 1M random examples and the two test sets contain 10K examples each.
We tokenize all inputs and targets by characters and decimal digits. 
We score the output by sequence-level accuracy, i.e., whether the output entirely matches the target sequence.
For convenience, we denote $\langle$EOS$\rangle$ as \texttt{\$}.

\paragraph{Number Sequence Prediction}
As the name suggests, the goal of the number sequence prediction problem \cite{nam2019number} is to predict the next term of an integer sequence.
After randomly choosing three initial terms, we generate a sequence via the recursion $a_n=2a_{n-1} - a_{n-2} + a_{n-3}$ which progresses the sequence up to the $n$\textsuperscript{th} term.
The input is the first $n$ terms $a_0, \dots , a_{n-1}$ and the target is the last term $a_n$.
The difficulty of the instance is parameterized by the maximum number of digits of the initial terms $a_0, \dots, a_{k-1}$, i.e., \textit{length}, and the total number of input integer terms $n$, i.e., \textit{$\#$terms}.
Those two difficulty parameters, \textit{length} and  \textit{$\#$terms}, vary (1-4, 4-6), (4, 4-6), and (6, 10-12) for the training set, the ID test set, and the OOD test set, respectively.
The input and the target of a training example are as follows.
\begin{flalign*}
    \text{  Input: }&\texttt{7008 -205 4 7221\$} &\\ 
    \text{  Target: }&\texttt{14233\$} & 
\end{flalign*}

\paragraph{Algebraic Word Problem}
To test the arithmetic abilities under linguistic instructions, we choose algebraic word problems, i.e., add-or-sub word,  \cite{saxton2019analysing}.
The difficulty of the problems is controlled by \textit{entropy}, the number of digits within a question.
Here, we make two differences from the original dataset.
First, we only allow integers whereas floating-point numbers can appear originally.
Second, our \textit{entropy} is the total number of digits in the input, whereas the original entropy is the maximum number of digits that input can have.
Our \textit{entropy} varies 16-20, 16-20, and 32-40 for the training set, the ID test set, and the OOD test set, respectively.
In the OOD test, we also impose every integer to be of length above 16 to guarantee that it is longer than any integers in the training set.
The input and the target of a training example are as follows. 
\begin{flalign*}
    \text{  Input: }&\texttt{What is -784518 take away 7323?\$}& \\
    \text{  Target: }&\texttt{-791841\$}&
\end{flalign*}

\paragraph{Computer Program Evaluation}
Predicting the execution results of programs requires algorithmic reasoning such as doing arithmetic operations or following programming instructions like variable assignments, branches, and loops.
We use \textit{mixed strategy} \cite{zaremba2014learning} to generate the training data with \textit{nesting} 2 and \textit{length} 5. 
For the ID test set and the OOD test set, \textit{nesting} and \textit{length} are set to be (2, 5) and (2, 7), respectively.
The input, a random Python snippet, and the target, the execution result, of a training example are as follows.
\begin{flalign*}
    \text{  Input: }&\texttt{j=891}\\
                    &\texttt{for x in range(11):j-=878}&\\
                    &\texttt{print((368 if 821<874 else j))\$}&\\
    \text{  Target: }&\texttt{368\$}&
\end{flalign*}

\subsection{Grid Decoder}
For solving arithmetic and algorithmic problems with digits, it is desirable to choose a grid decoder that can do local and parallel computation.
Therefore, we implemented a CNN \cite{he2016deep} consisting of three stacks of 3-layer bottleneck building blocks of ResNet.
Also, we implemented its attentional variant ACNN \cite{ramachandran2019stand}; every $3{\times}3$ convolution of the CNN was substituted with a stand-alone self-attention convolution.
We used $3{\times}25$-sized grids from the seq2grid module having 3-layered GRU encoder of hidden size 128 for both decoders.
As shown in Figure~\ref{fig:S2G-CNN}, we measured cross-entropy loss between the flipped-and-padded target and the output from the logit layer.
Here, the loss for empty symbol $\emptyset$ was included as we read out logits backward in the inference stage.
We jointly trained the seq2grid module and the CNN (ACNN) by the ADAM optimizer \cite{kingma2014adam} with a learning rate $1e^{-3}$.

\subsection{Results}
\begin{table}[t]
\centering
\resizebox{\columnwidth}{!}{
\begin{tabular}{lccccccccc}\toprule
& \multicolumn{2}{c}{Sequence} && \multicolumn{2}{c}{Add-or-sub} && \multicolumn{2}{c}{Program}\\
\cmidrule{2-3} \cmidrule{5-6} \cmidrule{8-9}
& ID & OOD && ID & OOD && ID & OOD\\ \midrule
Baselines\\
\phantom{a}LSTM        & 0.21 & 0.00 && 0.99 & 0.00 && 0.25 & 0.07 \\
\phantom{a}LSTM-Atten  & 0.68 & 0.00 && \textbf{1.00} & 0.00 && 0.37 & 0.01 \\
\phantom{a}RMC         & 0.01 & 0.00 && 0.99 & 0.00 && 0.33 & 0.01 \\
\phantom{a}Transformer & 0.97 & 0.00 && 0.97 & 0.00 && 0.37 & 0.00 \\
\phantom{a}UT          & \textbf{1.00} & 0.00 && \textbf{1.00} & 0.00 && \textbf{0.62} & 0.00 \\
\cmidrule{1-9}
Ours\\
\phantom{a}S2G-CNN     & 0.96 & \textbf{0.99} && 0.98  & 0.53 && 0.51 & 0.33 \\
\phantom{a}S2G-ACNN    & 0.90 & 0.92 && 0.96 & \textbf{0.55} && 0.44 & \textbf{0.35} \\
\bottomrule
\end{tabular}
}
\caption{Best sequence-level accuracy (out of 5 runs) on number sequence prediction problems (sequence), algebraic word problems (Add-or-sub), and computer program evaluation problems (Program) }
\label{table:arithmetic_problems_results}
\end{table}

\begin{figure}[t]
\begin{center}
\centerline{\includegraphics[width=\columnwidth]{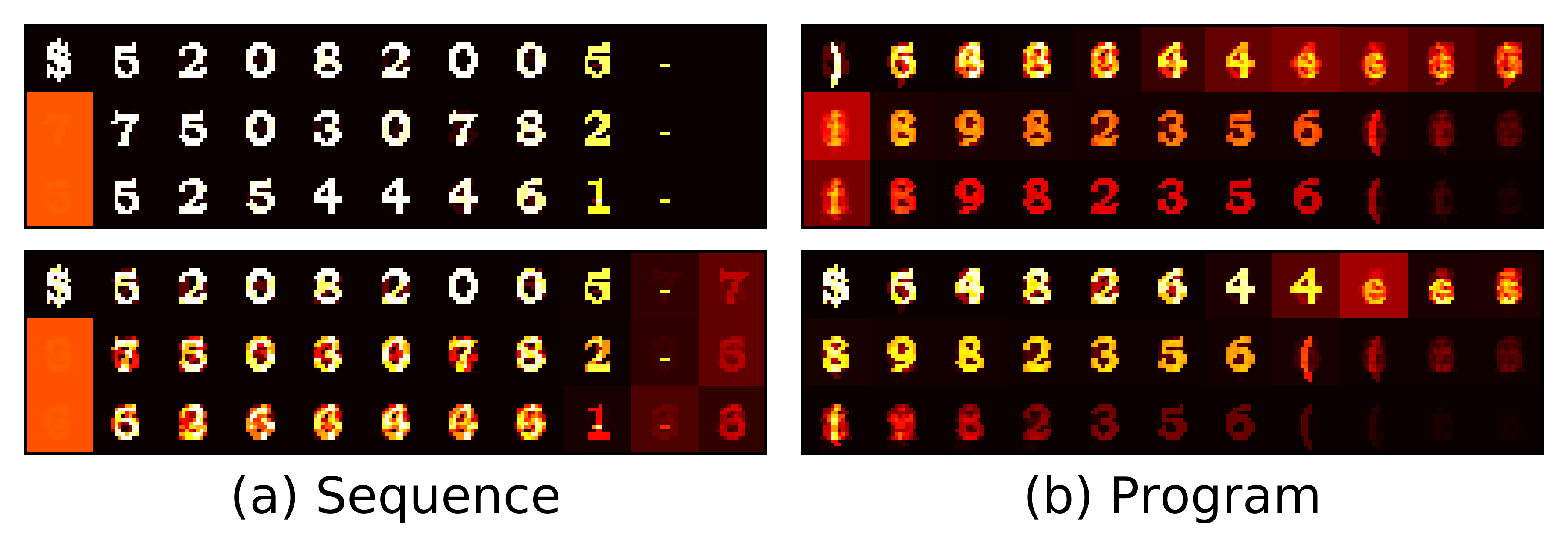}}
\caption{
Visualizations of preprocessed grid inputs of (a) number sequence prediction problems and (b) computer program evaluation problems.
The top and the bottom row correspond to S2G-CNN and S2G-ACNN, respectively.
}
\label{fig:grid_visualization}
\end{center}
\end{figure}

\begin{table}[]
\begin{center}
\begin{sc}
\begin{tabular}{lcccc}
\toprule
                      &  \normalfont{instruction}    & \normalfont{ID} & \normalfont{OOD}    \\ 
\midrule            
\multirow{3}{*}{\normalfont{LSTM-Atten}}      & \texttt{if-else}  & 0.46          & 0.26 \\
& \texttt{for}      & 0.06          & 0.03 \\
                                              & \texttt{*}        & 0.07          & 0.04 \\
                                              
\midrule
\multirow{3}{*}{\normalfont{UT}}              & \texttt{if-else}  & \textbf{0.81} & 0.01 \\
& \texttt{for}      & \textbf{0.38} & 0.00 \\
                                              & \texttt{*}        & \textbf{0.52} & 0.00 \\
                                              
\midrule
\multirow{3}{*}{\normalfont{S2G-CNN}}& \texttt{if-else}  & 0.73          & \textbf{0.57} \\                       & \texttt{for}      & 0.20          & 0.09 \\
                                              & \texttt{*}        & 0.25          & 0.14 \\
\bottomrule
\end{tabular}
\end{sc}
\end{center}
\caption{
Accuracy by instruction types of the best runs on the computer program evaluation problems.
For example, the S2G-CNN correctly answers 73\% of all ID snippets containing \texttt{IF-ELSE} instructions.
}
\label{table_program_by_instruction_result}
\end{table} \begin{figure}[t]
\begin{center}
\resizebox{1.0\columnwidth}{!}{
\begin{tabular}{l}
\toprule
\texttt{print((11*7288719))}\\
\midrule
\texttt{print(((6110039 if 7327755<3501784 else }\\
\texttt{1005398)*11))}\\
\midrule
\texttt{b=6367476}\\
\texttt{for x in range(19):b-=9082877}\\
\texttt{print((3569363 if 7448172<9420320 else b))}\\
\midrule
\texttt{e=(450693 if 4556818<2999168 else 3618338)}\\
\texttt{for x in range(10):e-=4489485}\\
\texttt{print(e)}\\
\bottomrule
\end{tabular}
} \caption{
Some OOD code snippets correctly answered by the best run of the S2G-CNN.
Note that snippets contain \texttt{FOR} or \texttt{*} instruction requiring non-linear time complexity.
}
\label{fig:S2G-ACNN_examples}
\end{center}
\end{figure}

Table~\ref{table:arithmetic_problems_results} shows that our models, S2G-CNN and S2G-ACNN, can generalize on OOD test sets.
In particular, both grid decoders achieve similar OOD generalization, implying that feeding the grid input via our seq2grid module can be beneficial to any decoder that can do local and parallel computations.
On the other hand, all baselines catastrophically fail at the OOD test sets although they seemingly perform well on the ID test set.
This shows that extending rules to longer numbers via sequential processing is extremely difficult.

As for the number sequence prediction problems, their OOD test results serve as unit tests for the seq2grid module since it needs to align digit symbols on the grid according to their scales.
Indeed, Figure~\ref{fig:grid_visualization}a shows that our module automatically finds such alignments that resemble the tailored grid of digits as shown in Figure~\ref{fig:toy_a}.

For the algebraic word problems, they require context-dependent arithmetic unlike number sequence prediction problems using the fixed progression rules.
In particular, linguistic instructions like \texttt{add} or \texttt{take away} indicates how to add/subtract given two numbers in a specific order.
Since our grid decoders apply the fixed convolutional filters over the grid, linguistic instructions must be reflected in the grid input beforehand for doing context-dependent arithmetic.
This shows that our seq2grid module can infuse the instruction information into the grid input.

For the computer program evaluation problems, predicting the output of a code snippet demands an understanding of algorithmic rules like branching mechanisms or for-loop given as programming instructions \texttt{IF-ELSE} or \texttt{FOR}.
Also, computing \texttt{*} operations has non-linear time complexity, unlike addition or subtraction.
Hence, we further investigate accuracy on snippets by those instructions as shown in Table~\ref{table_program_by_instruction_result}.
For the OOD snippets containing \texttt{IF-ELSE} instructions, our S2G-CNN achieves 57\% accuracy for them.
Considering that they can contain other instructions besides branching one as shown in Figure~\ref{fig:S2G-ACNN_examples}, the accuracy is fairly high.
For the non-linear operations, the S2G-ACNN shows little understanding compared with the UT on the ID test set.
However, the UT fails to extend rules of \texttt{FOR} and \texttt{*} instructions on the OOD test set while the S2G-CNN does so on some examples as shown in Figure~\ref{fig:S2G-ACNN_examples}.
These are surprising in that both the seq2grid module and the ACNN grid decoder do linear time computations in the input length.

\section{bAbI QA Tasks}
\label{section:babi}
Given as natural language with a small vocabulary of around 170, the bAbI QA tasks \cite{weston2015towards} test 20 types of simple reasoning abilities such as \textit{counting, induction, deduction,} and \textit{path-finding.}
A problem instance consists of a story, a question, and the answer. 
Here, the story contains supporting sentences about the answer and distractors that are irrelevant sentences to the answer.
\begin{figure}[t]
\begin{center}
\centerline{\includegraphics[width=0.90\columnwidth]{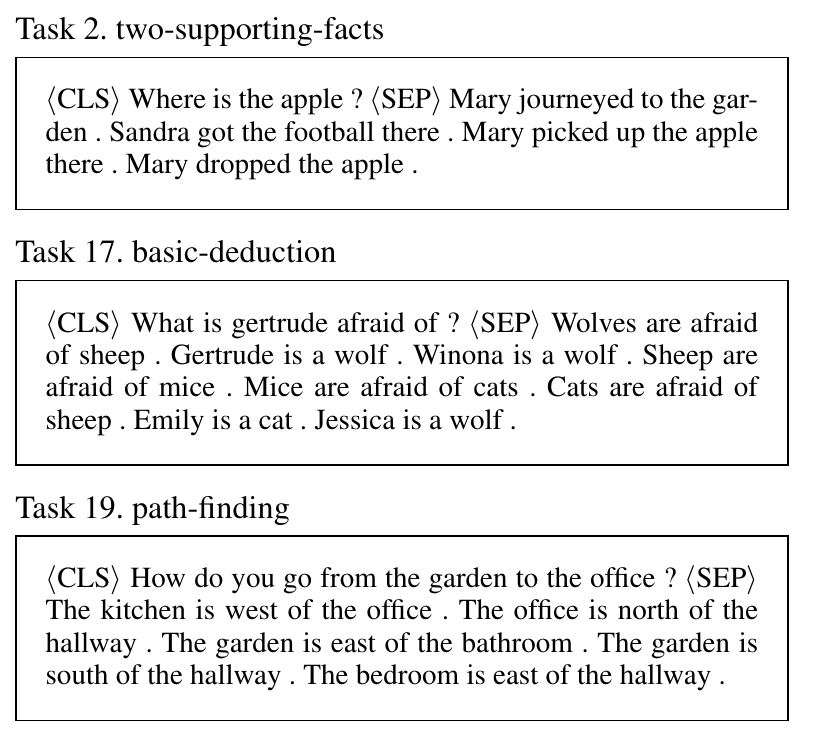}}
\caption{
Input examples of the bAbI QA tasks. }
\label{fig:babi_examples}
\end{center}
\end{figure}
We formulate the bAbI QA tasks \cite{weston2015towards} in a sequence classification setup such that an input is a concatenation of $\langle$CLS$\rangle$ token, a question, $\langle$SEP$\rangle$ token, and a story as shown in Figure~\ref{fig:babi_examples}.
While previous work \cite{dehghani2018universal} uses sentence-level encodings, we use straightforward one-hot word-level encodings.
This setup yields the increase of the average input length from 13.6 to 78.9, which in turn requires to handle much longer dependencies.
Hence, solving the bAbI tasks under word-level encodings is much harder than those under sentence-level encodings.
State-of-the-art models deal with longer dependencies via augmenting neural networks with external memory \cite{munkhdalai2019metalearned, rae2016scaling}.
However, we will show that the seq2grid module can enhance a simple neural network like TextCNN to effectively solve the word-level bAbI tasks, even in the absence of a complex and expensive memory structure.

\subsection{Grid Decoder}
\begin{figure}[t]
\centerline{\includegraphics[width=\columnwidth]{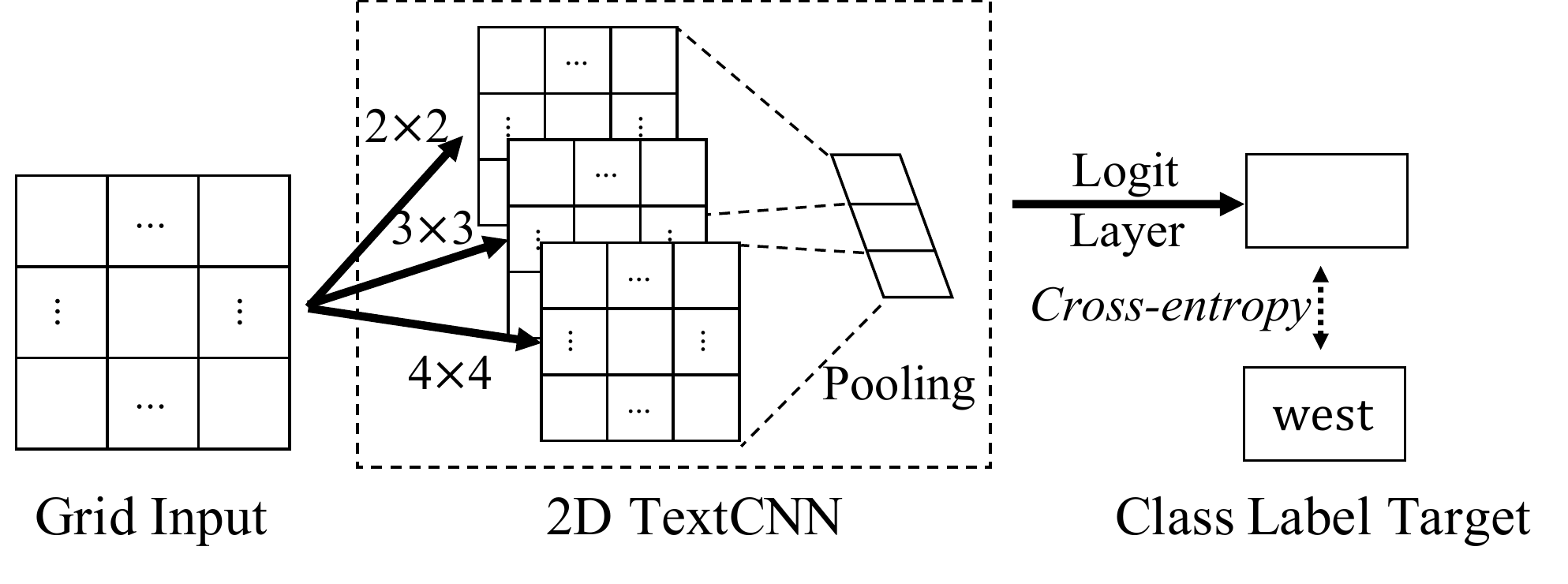}}
\caption{
The grid decoder of S2G-TextCNN.}
\label{fig:S2G-TextCNN}
\end{figure}

We chose a grid decoder as a variant of TextCNN \cite{kim2014convolutional}.
After the seq2grid module having 2-layered GRU encoders of hidden size 128 gave the $4{\times}8$-sized grid input, our TextCNN predicted the label by applying $k{\times}k$-CNNs ($k=2, 3, 4$), max-pooling, and dropout with the rate $0.4$ as shown in the Figure~\ref{fig:S2G-TextCNN}.
We used the ADAM optimizer to jointly train the seq2grid module and the TextCNN under a warm-up and decay learning rate scheme\textsuperscript{3}.

\begin{table}[t]
\centering
\resizebox{1.0\columnwidth}{!}{
\begin{tabular}{l@{~~~}l@{~~~}l@{~~~}l}\toprule
& $\#$params & Error & $\#$Failed tasks\\ 
\midrule
Baselines\protect\footnotemark\\
\phantom{a}LSTM & 25.6M & $24.9\pm 5.8$ & $12.1 \pm 3.7$\\
\phantom{a}Transformer & 0.5M & 	$33.1\pm1.7$	 & 	$18.9\pm0.3$\\
\phantom{a}UT & 0.5M &  	$26.8\pm6.0$	 & 	$15.0\pm4.0$\\
\phantom{a}TextCNN & 0.2M & $37.8\pm0.4$	 & 	$19.0\pm0.0$\\
\cmidrule{1-4}
Ours\\
\phantom{a}S2G-TextCNN & 0.8M & $10.8\pm0.8$ & 	$6.0\pm0.0$\\
\bottomrule
\end{tabular}
}
\caption{Error and $\#$Failed tasks ($> \! 5\%$ error) on the bAbI QA 10k joint tasks (for 10 runs).}
\label{table:babi_results}
\end{table}
\footnotetext{We use only encoders of baselines used in arithmetic and algorithmic problems.
As for the logit, LSTM uses the last hidden state while others use the hidden one corresponding to $\langle$CLS$\rangle$ token.}

\begin{table}[ht]\centering
\resizebox{1.0\columnwidth}{!}{
\def\arraystretch{1.1}
\begin{tabular}{@{}l@{}ccccc}
\toprule
& &\multicolumn{2}{c}{Baselines} && \multicolumn{1}{c}{Ours}\\ 
\cmidrule{3-4} \cmidrule{6-6}
\multirow{2}{*}{Task} & \multirow{2}{*}{\#supps} & \multirow{2}{*}{LSTM} & \multirow{2}{*}{UT} && S2G-\\
 &  & & && TextCNN\\
\midrule
1: single-supporting-fact & 1.0 & 0.0 & 0.0 & & 0.0 \\
2: two-supporting-facts & 2.0 & 47.4 & 55.0 & & 31.2 \\
3: three-supporting-facts & 3.0 & 45.9 & 67.9 & & 31.5 \\
4: two-arg-relations & 1.0 & 0.1 & 0.0 & & 0.0 \\
5: three-arg-relations & 1.0 & 0.8 & 5.5 & & 1.0 \\
6: yes-no-questions & 1.0 & 0.5 & 0.1 & & 0.0 \\
7: counting & 2.3 & 1.8 & 4.0 & & 0.0 \\
8: lists-sets & 1.9 & 0.2 & 2.3 & & 1.8 \\
9: simple-negation & 1.0 & 0.0 & 0.0 & & 0.0 \\
10: indefinite-knowledge & 1.0 & 0.3 & 0.0 & & 0.0 \\
11: basic-coreference & 2.0 & 0.0 & 0.1 & & 0.0 \\
12: conjunction & 1.0 & 0.0 & 0.0 & & 0.0 \\
13: compound-coreference & 2.0 & 0.0 & 0.0 & & 0.0 \\
14: time-reasoning & 2.0 & 20.6 & 4.4 & & 7.3 \\
15: basic-deduction & 2.0 & 34.8 & 18.5 & & 0.0 \\
16: basic-induction & 3.0 & 52.1 & 53.6 & & 51.7 \\
17: positional-reasoning & 2.0 & 41.1 & 41.0 & & 31.4 \\
18: size-reasoning & 2.0 & 8.6 & 9.1 & & 3.8 \\
19: path-finding & 2.0 & 90.9 & 79.1 & & 35.1 \\
20: agents-motivations & 1.0 & 1.8 & 1.4 & & 0.0 \\
\midrule
Mean error (\%) & & 17.3& 17.1&& 9.7  \\
\#Failed tasks&  &8 &8 && 6\\
\bottomrule
\end{tabular}
}
\caption{Task-wise errors on the bAbI QA 10k joint tasks for the best runs.
\#supps is the average number of supporting sentences in the story.
}
\label{table:babi_joint_best_results}
\end{table} 
\subsection{Results}
Our S2G-TextCNN outperforms sequential baseline models, such as the LSTM, the Transformer encoder, and the UT encoder, as shown in Table~\ref{table:babi_results}.
Note that we fed word-level inputs that require doing reasoning over distant symbols, i.e., the average length of inputs is 78.9, and we used the grid that has only 32 $(=\!4{\times}8)$ slots.
From these setups, we can conclude that our module can compress long inputs into grid inputs while selecting only necessary words along story arcs.
Moreover, the compression is effective in terms of the number of parameters.
Indeed, the GRU encoder inside our module is much smaller than the LSTM but enough to provide grid inputs to our grid decoder for solving the bAbI tasks.

We highlight that our seq2grid module, not the TextCNN decoder, leads to the superior performance of our model.
Since the attempt to use the usual TextCNN alone fails at almost all tasks, the dramatic performance gain by the aid of the seq2grid module is somewhat surprising.

We further analyze errors by tasks to see the possibility and the limitation of our sequence-to-grid method.
The zero variance in the number of failed tasks (Table~\ref{table:babi_results}) indicates that the S2G-TextCNN consistently fails on the same set of tasks, as listed in Table~\ref{table:babi_joint_best_results}.
Those failed tasks including \textit{two-supporting-facts, positional reasoning,} and \textit{path-finding} seem reasonably difficult for our models in that all of them require more than one supporting sentence for the reasoning.

\section{Conclusion}
\label{conclusion}
We introduced a neural sequence-to-grid (seq2grid) module which automatically segments and aligns an sequential input into a grid.
Our module was used as an input preprocessor for a neural network that took a grid input.
In particular, our module executed our novel nested list operations, ensuring an end-to-end joint training with the neural network.
Empirically, our module enhanced neural networks in various symbolic reasoning tasks.
 
\section{Acknowledgements}
\label{acknowledgements}
The authors appreciate Hyunkyung Bae for assistance with experiments. 
K. Jung is with ASRI and ECE, Seoul National University, Korea.
This work was supported by Samsung Research Funding $\&$ Incubation Center of Samsung Electronics under Project Number SRFCIT1902-06. 
\bibliography{references.bib}
\end{document}